%% file: 0.main.tex
\documentclass[10pt, a4paper]{article}

\usepackage[final]{lrec2026} 

\usepackage{times}
\usepackage{latexsym}
\usepackage{booktabs}
\usepackage{amsmath}
\usepackage{pifont}
\usepackage{graphicx}
\usepackage{import}
\usepackage{tcolorbox}
\usepackage{stmaryrd}
\usepackage{tikz}
\usetikzlibrary{positioning}

\usepackage{amsthm}

\newtheorem{definition}{Definition}

\allowdisplaybreaks

\usepackage[T1]{fontenc}

\usepackage[utf8]{inputenc}

\usepackage{microtype}

\usepackage{inconsolata}

\usepackage{graphicx}

\input{math_commands}

\newcommand{\rstat}[2]{$#1~\small{\pm\!#2}$}

\title{A Study on Building Efficient Zero-Shot Relation Extraction Models}

\name{Hugo Thomas \quad Caio Corro \quad Guillaume Gravier \quad Pascale Sébillot} 

\address{%
Univ Rennes, INSA Rennes, CNRS, Inria, IRISA -- UMR 6074, F-35042 Rennes, France
\\
\texttt{\{firstname.lastname\}@irisa.fr}}

\abstract{
Zero-shot relation extraction aims to identify relations between entity mentions using textual descriptions of novel types (i.e.,\ previously unseen) instead of labeled training examples.
Previous works often rely on unrealistic assumptions:
(1)~pairs of mentions are often encoded directly in the input, which prevents offline pre-computation for large scale document database querying;
(2)~no rejection mechanism is introduced, biasing the evaluation when using these models in a retrieval scenario where some (and often most) inputs are irrelevant and must be ignored.
In this work, we study the robustness of existing zero-shot relation extraction models when adapting them to a realistic extraction scenario.
To this end, we introduce a typology of existing models, and propose several strategies to build single pass models and models with a rejection mechanism.
We adapt several state-of-the-art tools, and compare them in this challenging setting,
showing that no existing work is really robust to realistic assumptions,
but overall \textsc{AlignRE} \citep{li2024alignre} performs best along all criteria.
\\ \newline
}

\begin{document}

\maketitleabstract

\input{1.introduction}

\input{2.related_works}

\input{3.typology}

\input{4.model}

\input{5.rejection_mechanism}
\input{6.conclusion}

\section*{Acknowledgments}
Experiments were carried out using the Grid'5000 testbed, supported by a scientific interest group hosted by Inria and including CNRS, RENATER and several universities (\url{https://www.grid5000.fr}). Hugo Thomas was funded by the "AI for semantic data analytics" doctoral program ANR-20-THIA-0018-01 and by Région Bretagne. Caio Corro is supported by the SEMIAMOR (CE23-2023-0005) and InExtenso (ANR-23-IAS1-0004) project grants of the French National Research Agency. This work contributes to the joint laboratory SYNAPSES (grant ANR-23-LCV2-0007) and to the AI Excellence Cluster SequoIA (grant ANR-23-IACL-0009).

\section*{Bibliographical References}
\bibliographystyle{lrec2026-natbib}
\bibliography{biblio}

\appendix

\end{document}

%% file: math_commands.tex
\usepackage{amsmath,amsfonts,bm}

\def\eqref#1{equation~\ref{#1}}

\def\1{\bm{1}}

\def\vs{{\bm{s}}}

\def\vu{{\bm{u}}}

\def\vw{{\bm{w}}}

\DeclareMathAlphabet{\mathsfit}{\encodingdefault}{\sfdefault}{m}{sl}
\SetMathAlphabet{\mathsfit}{bold}{\encodingdefault}{\sfdefault}{bx}{n}

\newcommand{\R}{\mathbb{R}}

\DeclareMathOperator*{\argmax}{arg\,max}

%% file: 1.introduction.tex
\section{\label{sec:intro}Introduction}

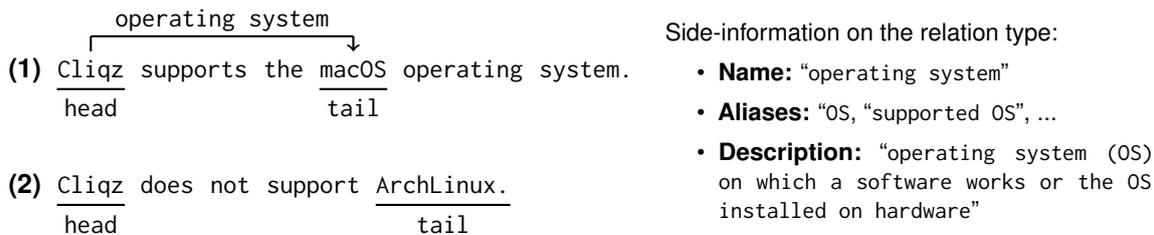
\begin{figure*}[t]
\null\hfill\begin{minipage}{0.51\textwidth}
    \input{figures/utterance_illustration}
\end{minipage}\hfill%
\begin{minipage}{0.4\textwidth}\small
    Side-information on the relation type:
    \begin{itemize}
        \item \textbf{Name:} ``\texttt{operating system}''
        \item \textbf{Aliases:} ``\texttt{OS}, ``\texttt{supported OS}'', ...
        \item \textbf{Description:} ``\texttt{operating system (OS) on which a software works or the OS installed on hardware}''
    \end{itemize}
\end{minipage}\hfill\null
    \caption{In relation extraction, the input is an utterance with two identified mentions, a head mention and a tail mention. We assume that the only targeted relation type is \texttt{operating system}.
    \textbf{(left)}
    Two input examples.
    In (1), the model must predict that there is a relation of this type between the two mentions.
    In (2), the model should reject the input, as the candidate input relation does not correspond to any type in the targeted ones.
    \textbf{(right)}
    Example of side-information used for zero-shot relation extraction.}
    \label{fig:utterance_and_description_illustration}
\end{figure*}

Relation extraction (RE) is a fundamental natural language processing task that aims to identify which relation links two entity mentions, if any---see Figure~\ref{fig:utterance_and_description_illustration} (left).
RE is central in many fields including biomedical research \citep{habibi2017deep}, finance \citep{hamad2024fire} and legal analysis \citep{zhong2020does}, mainly as an intermediate step for solving downstream tasks, including in information retrieval \citep{yang2024survey}, question answering \citep{chen2019uhop}, document retrieval for clinical decision process \citep{agosti2018relation}, etc.

Although several surveys on RE have been published \citep[][\emph{inter alia}]{hang2025few,zhao2024comprehensive,deng2024information,bassignana2022mean},
we argue that there is a need to better understand existing models in the context of specific use cases.
Indeed, users may face challenging settings that encompass constraints that have been  little studied in the literature.
In this work, we focus on zero-shot RE, which we cast as a relation mining problem where we assume the targeted text dataset is large but known in advance, e.g.,\ a company's document archive.
We illustrate this use case with the following scenario.

\begin{tcolorbox}[width=\linewidth, sharp corners=all, colback=white!95!black]
\paragraph{Use case scenario.}
A journalist wants to search for specific facts in a collection of raw news archives.
He starts by describing relation types of interest: ``\texttt{country in which this person rigged elections}''; ``\texttt{person who illegally financed elections in this country}''; etc.
Then, he uses a RE model to retrieve all occurrences of these relations in the archives.
\end{tcolorbox}

This setting is particularly challenging.
First, the targeted relation types are not known in advance, which we call \emph{on-the-fly} zero-shot RE, in order to insist that each novel query on the data may request for previously unknown types.
Second, to ensure reasonable processing times, models must allow to pre-compute and store input text representations.
In this work, we refer to this property as \emph{offline encoding}.

\begin{definition}[On-the-fly zero-shot classification]
    On-the-fly zero-shot classification refers to the problem of multiclass classification for which the set of output classes is (1)~unseen during training and (2) instance dependent, that this the set of output classes is specified at the same time as the instance to classify; in other words the set of output classes is part of the input.
\end{definition}
\begin{definition}[Offline encoding]
    Offline encoding refers to the setting where representations of instances (e.g., contextual embeddings of tokens for textual inputs) are pre-computed and stored in advance.
\end{definition}

These two requirements constrain the neural architecture to be based on \emph{late interaction} \cite{khattab2020colbert}, that is the encoding step (e.g.,\ using \textsc{Bert} or related models) must compute representations of the input utterance and the relation types separately, and only combine them when computing output logits, see Figure~\ref{fig:ZSRE_model_illustration}. 
In addition to late interaction, offline encoding means that entity mention candidates cannot be identified in the input during the utterance encoding step.

Last, when searching for relations of a given type in a large collection, most inputs should be rejected as they are irrelevant, and therefore the standard multi-class classification setting is inadequate.\footnote{For example, the second instance in Figure~\ref{fig:utterance_and_description_illustration} is rejected because the relationship between the two entity mentions does not correspond with to of the target relation types.}
This requires to augment the model with a \emph{rejection mechanism} \cite{hendrycks2016baseline,barandas2022uncertainty,hendrickx2024machine}.

\begin{tcolorbox}[width=\linewidth, sharp corners=all, colback=white!95!black]
\paragraph{Research question.}
Can we easily identify the best model for \emph{on-the-fly} and \emph{offline encoding} RE in large text collections,
and how robust are these models when enhanced with the necessary \emph{rejection mechanism} for this use-case?
\end{tcolorbox}

To answer this question, we propose a novel typology that shows that no recent off-the-shelf model is tailored to our use case.
Our classification shows that most approaches in the literature encode the targeted relation directly in the input by explicitly augmenting the input utterance to identify the head and tail mentions candidate of a relation: for instance, utterance (1) in Figure \ref{fig:utterance_and_description_illustration} can have its entities targeted using special markup: ``\texttt{<E1>Cliqz</E1> supports the <E2>macOS</E2> operating system.}''.
This means that these models do not enable offline encoding, even when they are based on a late-interaction architecture.\footnote{RE focuses on classifying a relation candidate identified by two mentions. In the zero-shot learning scenario, the targeted types may concern mention types that are unknown during the pre-computation stage. Note however that, as in previous zero-shot RE works, for evaluation we assume the gold mentions are given.}
Moreover, to the best of our knowledge, our typology shows that no recent zero-shot RE based on a late-interaction architecture implements a ``native'' rejection mechanism.

Using our typology, we identify three state-of-the-art models whose source codes are publicly available: \textsc{Emma} \cite{li2024fusion}, \textsc{ReMatching} \cite{zhao2023re} and \textsc{AlignRE} \cite{li2024alignre}.
We precisely describe the differences between these models, and explain how to adapt them to our use case, i.e.,\ to allow offline encoding and on-the-fly zero-shot RE.
As all these models lack a rejection mechanism, we present three different options based on previous works.
Finally, we evaluate these updated models with and without rejection mechanisms, on two publicly available datasets, \textsc{FewRel} \citep{han2018fewrel} and \textsc{WikiZSL} \citep{chen2021zs}.

Our experiments show that \textsc{AlignRE}, when adapted for offline encoding and a rejection mechanism, performs best among all considered models.

Our contributions can be summarized as follows:
(1)~we build a typology of zero-shot RE models;
(2)~we compare the main differences in the architecture of three state-of-the art models after adjustment for offline encoding;
(3)~we describe three rejection mechanisms that can be implemented in any zero-shot RE model;
(4)~we evaluate these models in comparable evaluation settings.
Code to reproduce experiments is publicly available.\footnote{\url{https://gitlab.inria.fr/huthomas/zsre-models-adaptation}}

%% file: figures/utterance_illustration.tex
\begin{tikzpicture}[
    every node/.style={
        rectangle,
        inner xsep=0cm,
        inner ysep=0.1cm,
        text height=1.5ex,
        text depth=.25ex,
    }
]
    %
    %
    
    \node (w10) [rectangle] {\textbf{(1)}};
    \node (w11) [rectangle, right=0.2cm of w10] {\texttt{Cliqz}};
    \node (w12) [rectangle, right=0.2cm of w11] {\texttt{supports}};
    \node (w13) [rectangle, right=0.2cm of w12] {\texttt{the}};
    \node (w14) [rectangle, right=0.2cm of w13] {\texttt{macOS}};
    \node (w15) [rectangle, right=0.2cm of w14] {\texttt{operating}};
    \node (w16) [rectangle, right=0.2cm of w15] {\texttt{system.}};

    \draw[thick] (w11.south west) -- coordinate (part1) node[yshift=-0.25cm] {\texttt{head}}  (w11.south east);
    \draw[thick] (w14.south west) -- coordinate (part2) node[yshift=-0.25cm] {\texttt{tail}}  (w14.south east);

    \coordinate[above=0.5cm of part1] (part1_a);
    \coordinate[above=0.7cm of part1] (part1_b);
    \draw[thick] (part1_a) -- (part1_b);
    \coordinate[above=0.5cm of part2] (part2_a);
    \coordinate[above=0.7cm of part2] (part2_b);
    \draw[thick,<-] (part2_a) -- (part2_b);
    \draw[thick] (part1_b) -- node[yshift=0.25cm] {\texttt{operating system}} (part2_b);

    %
    %
    
    \node (w20) [rectangle, below=1cm of w10] {\textbf{(2)}};
    \node (w21) [rectangle, right=0.2cm of w20] {\texttt{Cliqz}};
    \node (w22) [rectangle, right=0.2cm of w21] {\texttt{does}};
    \node (w23) [rectangle, right=0.2cm of w22] {\texttt{not}};
    \node (w24) [rectangle, right=0.2cm of w23] {\texttt{support}};
    \node (w25) [rectangle, right=0.2cm of w24] {\texttt{ArchLinux.}};

    \draw[thick] (w21.south west) -- node[yshift=-0.25cm] {\texttt{head}} (w21.south east);
    \draw[thick] (w25.south west) -- node[yshift=-0.25cm] {\texttt{tail}} (w25.south east);
\end{tikzpicture}

%% file: 2.related_works.tex
\section{Related Works}
\label{sec:related_work}

In this section, we review previous publications following similar goals as ours,
giving a clear view over the current state of models and identifying design choices that impact downstream results.

\subsection{Related Typologies}

Building a typology, i.e.,\ a set of targeted criteria, enables systematic model comparison and selection for specific scenarios.

\citet{zhao2024comprehensive} focus on \emph{supervised} RE, whereas we focus on zero-shot learning.
Closer to our setting, \citet{hang2025few} introduced a taxonomy of manual and automatic prompts for few-shot RE.
They analyze template construction and model fine-tuning strategies, as well as their pros and cons (e.g.,\ annotation costs, prediction time, etc).
We focus on other criteria, namely processing efficiency and rejection methods.

Beyond RE,
\citet{deng2024information} and \citet{pai-etal-2024-survey} studied low-resource and open information extraction, respectively, but do not consider zero-shot RE.
We take inspiration from these studies, differentiating ourselves by the goal of our typology, that is finding adapted or adaptable models to large scale and on-the-fly relation extraction thanks to offline encoding.

\subsection{Datasets and Evaluation}

\citet{bassignana2022mean} proposed a broad overview on RE datasets and evaluation protocols.
Importantly, they show that annotation is not consistent across datasets.
However, their analysis does not mention the presence or absence of side-information about relation types (like textual description, aliases, etc.) that are mandatory for our zero-shot setting.

\citet{han2018fewrel} introduced the \textsc{FewRel} dataset for few-shot RE.
Relation types and their instances are based on Wikidata.
\citet{chen2021zs} proposed to use the same dataset for zero-shot RE by simply changing the train/test splits.
However, this dataset is tailored for the \emph{classification} scenario,
that is each sample of the test split is guaranteed to belong to a fixed set of relation types. 
Datasets \textsc{ReTACRED} \citep{stoica2020re} and \textsc{Nyt} \citep{riedel2010nyt} include a special relation type for rejection evaluation,
i.e.,\ a specific output class that identifies couples of mentions that do not belong to the targeted relation types.
However, these datasets do not include side-information about relation types, and can therefore not be straightforwardly used for zero-shot learning.

Unfortunately, to the best of our knowledge, no currently available RE dataset was initially built including both data for rejection evaluation and side-information required for zero-shot learning, except \textsc{FewRel~2.0} \citep{gao2019fewrel}.
However, its authors' goal is to identify relations between mentions with respect to the whole set of types,
whereas in our RE scenario,
we aim to reject a candidate with respect to an ``on-the-fly defined'' list of targeted relations.
Therefore, in this work, we choose to focus on \textsc{FewRel} and \textsc{WikiZSL}, with tailored evaluation procedures for rejection.
This choice is motivated by the fact that we want to study \emph{model robustness to adaptation for large scale and on-the-fly RE}: As such, we use the same dataset for evaluation with and without a rejection mechanism.
These datasets remain relatively small and well-annotated compared to the data we will encounter in our scenario. We justify this choice by the need for comparability to previous models and by the ease of evaluation through reliable and exhaustive annotation at the sentence level.

\subsection{Benchmarking Design Choices}

\citet{soares2019matching} and \citet{gravier2023derriere} compared different entity pair embedding methods, and conclude that surrounding the mentions of entity pairs with
markup and concatenating the markup's embeddings offers the best performance for RE, unfortunately preventing offline encoding.
This was also observed in ablation studies in \citet{zhao2023re,li2024alignre}.

In the biomedical domain,
\citet{sarrouti2022comparing} and \citet{naguib2024fewshot} compared encoder-only and encoder-decoder architectures for RE and named entity recognition, respectively.
They both conclude that encoder-only architectures lead to better results while having fewer parameters.

%% file: 3.typology.tex
\begin{table*}[t]
    \centering
    \input{results/typology2}
    \caption{Typology of zero-shot relation extraction models, ordered chronologically.
    Row \textsc{SBert} indicates the use of \textsc{SBert} to encode side-information of relation types.
    The F1 row contains results as reported by original authors on \textsc{FewRel} using 5 unknown relation types, without rejection mechanism. (We do not include results for \textsc{MatchPrompt} and \textsc{AlignRE} as they have not been tested on this dataset) Note that there is also a late interaction variant of \textsc{Emma}, but it achieves significantly lower F1 scores.}
    \label{tab:models}
\end{table*}

\begin{figure}
    \centering
    \includegraphics[width=0.45\textwidth]{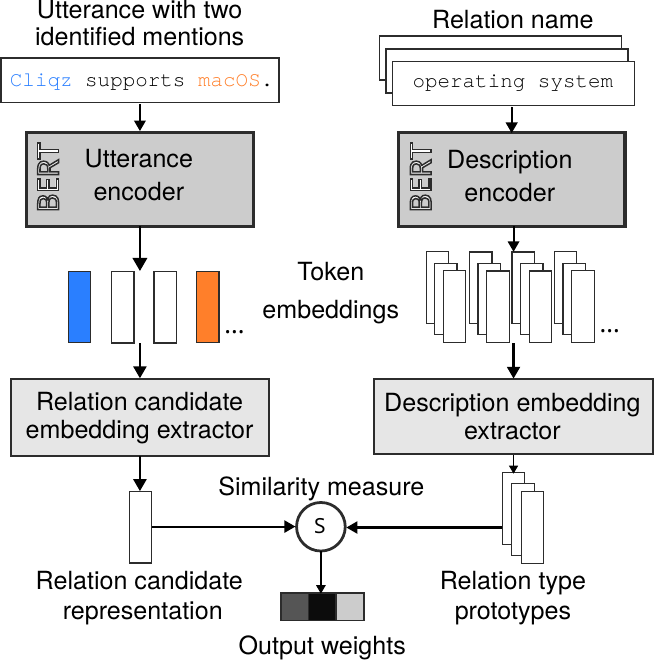}
    \caption{Generic illustration of an encoder-only zero-shot RE model.
    }
    \label{fig:ZSRE_model_illustration}
\end{figure}

\section{Typology of Zero-Shot RE Models}
\label{sec:typology}

We introduce a novel typology, summarized in Table~\ref{tab:models}, that allows to compare existing models for on-the-fly zero-shot RE with offline encoding, including the need for a rejection mechanism.\footnote{
Recent works also studied generative language models (LM) for zero-shot RE \citep[][\emph{inter alia}]{chia2022relationprompt,li2023revisiting,liu2024unleashing,zhang2023aligning}.
These approaches (including more recent retrieval-augmented generation approaches) need to run a full forward pass on the entire LM for each input utterance,
meaning that it is not possible to pre-compute and cache input representations for offline encoding.
We do not consider aforementioned models in this work
and focus solely on methods based on encoder-only architectures.}

\subsection{Encoders}

A simplified illustration of a generic encoder-only neural architecture as used in few-shot and zero-shot learning is depicted in Figure \ref{fig:ZSRE_model_illustration}: the neural network builds (1)~a representation of the relation candidate between a head entity and a tail entity in an input utterance using the \emph{utterance encoder} on the left side and (2)~a representation of each targeted relation type, also called a prototype, from their respective side-information using the \emph{description encoder} on the right side.
To this end, relation mentions and descriptions tokens are contextually encoded into token embeddings, which are processed into a single vector of a relation mention or relation type (prototype).
Then, prediction reduces to searching the best fit between the candidate representation and relation type prototypes, often by maximizing the cosine similarity.

We now describe the encoder properties highlighted by our typology.
First, although all models rely on \textsc{Bert} \citep{devlin-etal-2019-bert} or a similarly pre-trained encoder for the utterance encoder,
they differ in the description encoder:
some use the same network; others instead rely on the sentence embedding model of \citet{reimers-gurevych-2019-sentence}, which we call \textsc{SBert} in the following.

Second,
the utterance and description encoders can be joined into a single encoder (e.g.\ including cross-attention between the input utterance and relation type side-information) or they may be independent.
We say that independent encoders allow for \emph{late interaction}, which is mandatory for on-the-fly zero-shot RE with offline encoding.

\paragraph{Natural Language Inference (NLI).}
There is a single exception to this generic architecture: \textsc{LaVeEntail} \citep{sainz2021label}, for label verbalization and entailment, which is based on a NLI model.
The model takes as input the utterance (first sentence) and a relation type side-information (second sentence),
and predicts if the two sentences are related by entailment, disregarding in this case contradiction or neutrality. High entailment probability means the input relation candidate is of this relation type.

\subsection{Relation Type Representations}

The following side-information about relation types has been used to build relation type representations:
\begin{enumerate}
    \item \emph{name}, e.g.\ ``\texttt{operating system};
    \item \emph{description}, e.g.\ ``\texttt{operating system (OS) on which a software...}'';
    \item aliases, e.g.\ ``\texttt{OS}'', ``\texttt{supported OS}'';
    \item expected head and tail mention type names, e.g.\ ``\texttt{software}''.
\end{enumerate}
For the simplest case, \citet{boylan2025glirel} rely only on relation type names as a very short and dense source of information.
Instead, \citet{li2024fusion} choose to rely only on descriptions, whereas more involved approaches include name, description and aliases \citep{li2024alignre}.
Finally, \citet{zhao2023re} also incorporate head and tail mention type information.

\subsection{Relation Candidate Representation}
\label{sec:utterance_encoder}

Most few- and zero-shot RE models encode the relation candidate in the input of the utterance encoder.
For example, the first utterance and relation candidate in Figure~\ref{fig:utterance_and_description_illustration} (left) can be encoded as ``\texttt{<E1>Cliqz</E1> supports the <E2>macOS</E2> operating system.}''.

Such encoding breaks compatibility with our on-the-fly zero-shot setting.
Indeed, in the offline encoding step (i.e.,\ storing utterance representations of the whole targeted text database in advance),
we do not know what mention types will be of interest to the user; therefore, we cannot pre-identify relation candidates at this stage (as head and tail mention types depend on targeted relation types).

Therefore, an important property we seek for the utterance encoder is that it is \emph{single pass}: it takes as input the raw utterance ``\texttt{Cliqz supports the macOS operating system.}'',
and the same token representations of this utterance can be used no matter the target head and tail mentions, which allows on-the-fly zero-shot RE with offline encoding.

\subsection{Rejection Mechanism}

\textsc{LaVeEntail} \citep{sainz2021label} is the only model that includes a native rejection mechanism, based on an extra relation type for rejection whose description is ``\texttt{E1 and E2 are not related}'', where \texttt{E1} and \texttt{E2} are replaced head and tail mentions, respectively.
If this rejection description obtains the highest entailment score among all descriptions for an utterance, said utterance is rejected.\footnote{There are older works such as \citep{levy2017zero} that include a rejection mechanism, but they are not included in our typology as their experimental results are far from current state-of-the-art.}

%% file: results/typology2.tex
\scriptsize
\begin{tabular}{@{}l@{}c@{\hskip 0.5em}c@{\hskip 0.5em}c@{\hskip 0.5em}c@{\hskip 0.5em}c@{\hskip 0.5em}c@{\hskip 0.5em}c@{\hskip 0.5em}c@{\hskip 0.5em}c@{\hskip 0.5em}c@{}}
\toprule
& \textsc{Zs-Bert}
& \textsc{LaVeEntail}
& \textsc{Rcl}
& \textsc{MatchPrompt}
& \textsc{ReMatching}
& \textsc{Cl\&Cd}
& \textsc{Emma}
& \textsc{AlignRE}
& \textsc{Ce-Da}
& \textsc{GliRel}
\\
& {\scriptsize \citeauthor{chen2021zs}}
& {\scriptsize{} \citeauthor{sainz2021label}}
& {\scriptsize \citeauthor{wang2022rcl}}
& {\scriptsize{} \citeauthor{wang2022matchprompt}}
& {\scriptsize \citeauthor{zhao2023re}}
& {\scriptsize{} \citeauthor{yang2024cl}}
& {\scriptsize \citeauthor{li2024fusion}}
& {\scriptsize{} \citeauthor{li2024alignre}}
& {\scriptsize \citeauthor{zhang2025custom}}
& {\scriptsize{} \citeauthor{boylan2025glirel}}
\\
& {\scriptsize{} (\citeyear{chen2021zs})}
& {\scriptsize{} (\citeyear{sainz2021label})}
& {\scriptsize{} (\citeyear{wang2022rcl})}
& {\scriptsize{} (\citeyear{wang2022matchprompt})}
& {\scriptsize{} (\citeyear{zhao2023re})}
& {\scriptsize{} (\citeyear{yang2024cl})}
& {\scriptsize{} (\citeyear{li2024fusion})}
& {\scriptsize{} (\citeyear{li2024alignre})}
& {\scriptsize{} (\citeyear{zhang2025custom})}
& {\scriptsize{} (\citeyear{boylan2025glirel})}
\\
\midrule
\multicolumn{10}{l}{\textbf{Neural architecture}}\\
\midrule
\textsc{SBert}
& \ding{51}
& 
& 
& 
& \ding{51}
& 
& 
& \ding{51}
& 
& 
\\
Single pass
& \ding{51}
& 
& 
& 
& 
& 
& 
& 
& 
& \ding{51}
\\
Rej. mechanism
& 
& \ding{51}
& 
& 
& 
& 
& 
& 
& 
& 
\\
Late interaction
& \ding{51}
& 
& \ding{51}
& \ding{51}
& \ding{51}
& \ding{51}
& 
& \ding{51}
& \ding{51}
& 
\\
\midrule
\multicolumn{10}{l}{\textbf{Side-information}}\\
\midrule
Name
& 
& 
& 
& 
& 
& 
& 
& \ding{51}
& \ding{51}
& \ding{51}
\\
Description
& \ding{51}
& 
& 
& \ding{51}
& \ding{51}
& \ding{51}
& \ding{51}
& \ding{51}
& \ding{51}
& 
\\
Aliases
& 
& 
& 
& 
& 
& 
& 
& \ding{51}
& \ding{51}
& 
\\
Mention types
& 
& 
& 
& 
& \ding{51}
& 
& 
& 
& 
& 
\\
\midrule
\multicolumn{10}{l}{\textbf{Experimental Results}}\\
\midrule
Public code
& \ding{51}
& \ding{51}
& \ding{51}
& 
& \ding{51}
& 
& \ding{51}
& \ding{51}
& 
& \ding{51}
\\
F1
& 77.90
& 
& 90.73
& 
& 92.58
& 96.79
& 94.67
& 93.09
& 95.17
& 94.20
\\
\bottomrule
\end{tabular}

%% file: 4.model.tex
\section{Efficient Relation Extraction\\without Rejection}
\label{sec:efficient}

Next, we explain how we adapt each considered model to our relation extraction scenario.

\subsection{Selected Baselines}

Table~\ref{tab:models} shows that no model is tailored for our use case:
there is only one model which is single pass, but it results in way lower F1 score than others, and none of the late interaction model includes a rejection mechanism.

For this study, we select two state-of-the art late interaction models whose code is publicly available: \textsc{ReMatching} \citep{zhao2023re} and \textsc{AlignRe} \citep{li2024alignre}.
Moreover, we also include the variant without late interaction of \textsc{Emma} in the study, in order to understand how much performance drop when using late interaction.
We could have included \textsc{GliRel} instead, which also features early interaction, but we stuck with \textsc{EMMA} for its close resemblance to the two other models' architecture, which allows better comparison. 

For the three models,
we compare their \emph{off-the-shelf} implementations, that is software as distributed by authors that we train ourselves,
with our own adaptation that enables single pass inference.

\subsection{Single Pass Adaptation}

As described in Section~\ref{sec:utterance_encoder},
most models encode the relation candidate in the input utterance.
In this section, we briefly describe our single pass variants of the chosen baselines.
The main idea is that we give to the encoder the raw input utterance, and then use \textsc{Bert}'s output contextual embeddings as representation of targeted head and tail entity mentions.
With this approach, the whole unaltered input utterance is encoded, and at test-time any two mentions can be identified as relation head and tail candidates, by extracting their respective token embeddings.
For the three baselines, we update the model so that the single pass variant is as similar as possible to the original implementation.

Let $(\vs, e_1, e_2)$ be an input, where $\vs$ is an input sentence, and $e_1$ (resp.\ $e_2$) is the span of the head (resp.\ tail) mention of the relation.
We denote $d \in \mathbb N_{>0}$ the dimension of \textsc{Bert} outputs.

\paragraph{\textsc{Emma}.}
We first build a representation $f_\text{me.}(\vs, e_1)$ (resp.\ $f_\text{me.}(\vs, e_2)$) for the head (resp.\ tail) mention, and $f_\text{sent.}(\vs)$ for the global utterance.\footnote{Which is simply the contextual embedding of the \texttt{[CLS]} token.}
We test different strategies to build the representation of $e_1$ (similarly $e_2$), which are all based on the \textsc{BERT} model outputs, i.e.,\ contextual embeddings:
\begin{itemize}
      \item \textbf{first}: use the mention's first token embedding;
      \item \textbf{projection}: concatenate the first and last token embeddings of the mention, and then project this vector into dimension $d$;\footnote{Parameters of the projection are learned, and we use the same projection for $e_1$ and $e_2$.}
      \item \textbf{mean pooling}: average the first and last token embeddings of the mention;
      \item \textbf{max pooling}: compute element-wise maximum between first and last token embeddings.
\end{itemize}
Then, we build a relation candidate representation $f_\text{rel.}(\vs, e_1, e_2) \in \R^{3d}$
as follows:
\[
f_\text{rel.}(\vs, e_1, e_2)
=
f_\text{sent.}(\vs)
\oplus f_\text{me.}(\vs, e_1)
\oplus f_\text{me.}(\vs, e_2),
\]
where $\oplus$ denotes vector concatenation.
The rest of the model follows the original implementation.

\paragraph{\textsc{AlignRe}.}
For this model, the relation candidate representation $f_\text{rel.}(\vs, e_1, e_2) \in \R^d$ is the mean of head and tail mention representations:
\[
f_\text{rel.}(\vs, e_1, e_2)
=
\frac{1}{2}(
f_\text{me.}(\vs, e_1)
+ f_\text{me.}(\vs, e_2)),
\]
and we test the same strategies as for $\textsc{Emma}$.

When trying our different strategies on \textsc{AlignRE}, we remove the custom prompt that the original model appends when using entity markers, i.e.,\ ``\texttt{The relation between [MASK] E1 and [MASK] E2 is [MASK]}'', where E1 and E2 are replaced by the head and tail mentions, respectively.\footnote{The original model would build relation representations by concatenating the three [MASK] tokens' representations from the prompt, whereas we use the previously described strategies.}
Indeed, this extra input is not compatible with offline encoding as it pre-identifies targeted head and tail mentions.
The rest of the model follows the original implementation.

\paragraph{\textsc{ReMatching}.}
The upgrade is similar to the one of \textsc{Emma}, but instead of concatenating the 3 representations to build $f_\text{rel.}(\vs, e_1, e_2)$,
they are kept separated to compute 3 cosine similarities that are then aggregated, as per original implementation.

\subsection{Experiments}

\paragraph{Prediction.}
Let $T$ be the set of all relations annotated in a given dataset.
Let $T' \subseteq T$ be a subset of relation types.
Each model first computes a vector $\vw \in \R^{T'}$
Then, the prediction $\widehat t$ is simply the relation type of maximum score:
where $w_t$ is output score associated with relation type $t \in T'$ for the input.\footnote{$\vw \in \R^{T'}$ denotes the real-valued vectors indexed by elements of $T'$.}
\[
\widehat t = \argmax_{t \in T'} w_t.
\]

\paragraph{Data.}
We evaluate on \textsc{FewRel} \citep{han2018fewrel} and \textsc{WikiZSL} \citep{chen2021zs}.
Dataset statistics are given in Table~\ref{tab:datasets}.

\begin{table}[t]
    \centering
    \small
    \begin{tabular}{lcccc}
        \toprule
        & \textbf{|T|}& \textbf{|D|} & \textbf{Dist.\ me.}  & \textbf{Avg.\ len.} \\
        \midrule
        \textsc{WikiZSL}   & 113 & 94\,383    & 77\,623       & 24.85           \\
        \textsc{FewRel}    & 80  & 56\,000    & 72\,954        & 24.95           \\
        \bottomrule
    \end{tabular}
    \caption{Datasets statistics, where $T$ is the set of relation types, $D$ the set of relation quadruples including utterance, entities and relation type $(s, e_1, e_2, t)$, and the last two columns give the number of distinct entity mentions and the average length of utterances, respectively.}
    \label{tab:datasets}
\end{table}

\paragraph{Hyperparameters.}
Models are trained for 5 epochs with a learning rate of $2 \times 10^{-5}$ for \textsc{Emma} and $10^{-5}$ for \textsc{AlignRE} and \textsc{RE-Matching}, using \textsc{AdamW} \cite{loshchilov2018decoupled}.
For \textsc{AlignRE} and \textsc{RE-Matching}, the \textsc{SBert} encoder is frozen.
To obtain reliable results, all experiments are repeated 3 times with different random seeds, and we report average and standard deviation.

\paragraph{Evaluation metric.}
Note that in our evaluation setting,
an input is a tuple $(\vs, e_1, e_2)$,
and an output is a relation type $t \in T_\text{eval}$, where $T_\text{eval}$ is the evaluation set of relation types, unseen during training.
As such, this evaluation setting reduces to a standard multi-classification setting,
and we simply report the F1 score on the test set.

To better analyze robustness, we report F1 scores with different number of unknown classes in $T_\text{eval}$;
in practice we set $|T_\text{eval}| \in \{5, 10, 15\}$.
For a given set $T_\text{eval}$, the dataset is trivially split between train and evaluation depending on the gold annotated output for an input triple $(\vs, e_1, e_2)$.

\begin{table*}[h]
\small
\centering
\input{results/classification.tex}
\caption{Results in terms of macro F1, without rejection mechanism. We compare the off-the-shelf softwares with our efficient variants that allow on-the-fly zero-shot relation extraction with offline encoding.}
\label{tab:classification_results}
\end{table*}

\paragraph{Analysis.}
Results are given in Table~\ref{tab:classification_results}.
Strategies like mean pooling, max pooling or first obtain lower scores by a few points in some cases, but models \textsc{ReMatching} or \textsc{AlignRE} are able to mitigate this difference in performance: while still worse than \textsc{Emma} with 5 unseen types, they are better with more unseen relation types and remain stable when substituting the entity markup by other embedding strategies.
This shows that in this setting, late encoding does not necessarily hurt performances.
Overall, the efficient \textsc{ReMatching} variants perform best compared to other efficient variants,
but \textsc{AlignRE} remains competitive.

The concatenation of both first and last mention tokens embeddings results in the lowest scores.
\textsc{Emma} is overall the most affected by the change of embedding strategy, going from the highest scores to scores comparable or worse to other models.

Overall, the evaluated single pass strategies yield acceptable scores with minimal degradation compared to the original approaches, and are therefore suitable to our use case.

%% file: results/classification.tex
\begin{tabular}{@{}l@{\hskip 4em}ccc@{\hskip 4em}ccc@{}}
\toprule
&\multicolumn{3}{@{}c@{\hskip 4em}}{\textsc{FewRel}}
&\multicolumn{3}{@{}c@{}}{\textsc{WikiZSL}}
\\
\cmidrule(lr{4em}){2-4}
\cmidrule{5-7}
& 5 & 10 & 15
& 5 & 10 & 15
\\
\midrule
\multicolumn{7}{l}{\textsc{Emma}}
\\
\midrule
Off-the-shelf
& \rstat{98.4}{1.2} & \rstat{84.5}{4.5} & \rstat{79.5}{6.2} 
& \rstat{88.2}{7.0} & \rstat{67.9}{8.8} & \rstat{62.8}{6.6}
\\[0.3em]
\multicolumn{7}{@{}l@{}}{Our efficient variants}
\\
$\hookrightarrow$ first
& \rstat{94.5}{1.6}& \rstat{79.5}{3.0}& \rstat{70.2}{4.6}
& \rstat{81.6}{4.8} & \rstat{64.7}{7.7} & \rstat{57.0}{2.0}
\\
$\hookrightarrow$ projection
&\rstat{94.6}{1.2} & \rstat{77.8}{4.8} & \rstat{67.2}{6.1}
& \rstat{79.0}{6.2} & \rstat{59.0}{9.9} & \rstat{50.5}{2.3}
\\
$\hookrightarrow$ max pooling
& \rstat{94.6}{1.5}& \rstat{79.9}{2.3}& \rstat{70.2}{4.8}
& \rstat{83.9}{3.5} & \rstat{65.9}{9.1} & \rstat{58.4}{1.8}
\\
$\hookrightarrow$ mean pooling
& \rstat{94.6}{1.4} & \rstat{80.0}{2.5} & \rstat{70.9}{5.3} 
& \rstat{82.6}{3.4} & \rstat{65.5}{8.9} & \rstat{57.8}{3.7}
\\
\midrule
\multicolumn{7}{l}{\textsc{Re-Matching}}
\\
\midrule
Off-the-shelf
& \rstat{92.7}{4.1}& \rstat{84.1}{5.3} & \rstat{75.0}{3.7} 
& \rstat{86.3}{6.3} & \rstat{84.0}{6.3} & \rstat{74.3}{8.0}
\\[0.3em]
\multicolumn{7}{@{}l@{}}{Our efficient variants}
\\
$\hookrightarrow$ first
& \rstat{91.6}{4.7}& \rstat{84.0}{6.6}& \rstat{75.4}{3.8}
& \rstat{86.7}{6.1} & \rstat{85.2}{6.7} & \rstat{74.8}{8.5}
\\
$\hookrightarrow$ projection
& \rstat{77.7}{5.2} & \rstat{71.7}{2.8} & \rstat{59.8}{13.2}
& \rstat{73.0}{4.5} & \rstat{70.1}{5.8} & \rstat{55.6}{4.5}
\\
$\hookrightarrow$ max pooling
& \rstat{92.6}{3.9}& \rstat{84.2}{5.4}& \rstat{75.3}{3.8}
& \rstat{86.2}{6.0} & \rstat{84.2}{6.1} & \rstat{74.0}{8.5}
\\
$\hookrightarrow$ mean pooling
& \rstat{91.9}{4.4} & \rstat{83.7}{6.1} & \rstat{75.3}{4.8} 
& \rstat{86.8}{6.0} & \rstat{85.1}{6.6} & \rstat{75.1}{8.7}
\\
\midrule
\multicolumn{7}{l}{\textsc{AlignRE}}
\\
\midrule
Off-the-shelf
& \rstat{90.7}{0.2} & \rstat{84.9}{1.8} & \rstat{73.4}{5.1}
& \rstat{78.2}{4.3} & \rstat{73.8}{6.0} & \rstat{65.2}{7.7}
\\[0.3em]
\multicolumn{7}{@{}l@{}}{Our efficient variants}
\\
$\hookrightarrow$ first
& \rstat{89.5}{4.4}& \rstat{86.9}{0.8} & \rstat{72.7}{4.0}
& \rstat{80.1}{4.1} & \rstat{74.1}{2.9} & \rstat{68.1}{4.2}
\\
$\hookrightarrow$ projection
& \rstat{51.4}{10.9} & \rstat{55.2}{5.4} & \rstat{39.7}{4.7}
& \rstat{76.7}{6.5} & \rstat{44.3}{3.2} & \rstat{46.5}{4.6}
\\
$\hookrightarrow$ max pooling
& \rstat{90.0}{3.5}& \rstat{87.6}{1.3}& \rstat{73.5}{4.7} 
& \rstat{77.9}{6.5} & \rstat{72.4}{3.7} & \rstat{67.4}{4.7}
\\
$\hookrightarrow$ mean pooling
& \rstat{90.4}{2.9} & \rstat{87.9}{3.4} & \rstat{73.0}{3.3} 
& \rstat{80.6}{4.0} & \rstat{74.3}{4.5} & \rstat{67.9}{3.2}
\\
\bottomrule
\end{tabular}

%% file: 5.rejection_mechanism.tex
\section{Rejection Mechanism}
\label{sec:rejection}

In the previous section, we compared off-the-shelf versions of three zero-shot RE models with our updated efficient variants.
Unfortunately, this evaluation setting is quite artificial as it assumes that each relation candidate is known to be of one of the targeted types  $T_\text{eval}$.
In practice, we often aim for relation \emph{extraction}, where, for each relation candidate, we must decide if it belongs to a type in $T_\text{eval}$ or not.
In other words, we must have an option to \emph{reject} a candidate.

To this end, we assume an extra set of relation types $R$ such that $R \cap T = \emptyset$,
where all relation types in $R$ are considered as \emph{reject types}.
For a given input $(\vs, e_1, e_2)$,
the model now builds an augmented score vector $\overline{\vw} \in \R^{T'\cup R}$, $T' \subseteq T$
s.t.:
\[
\overline{w}_t = \begin{cases}
    w_t \quad&\text{if~} t \in T',\\
    u_t \quad&\text{otherwise,}
\end{cases}
\]
where $\vw$ is built as described in Section~\ref{sec:efficient},
and $\vu \in \R^R$ is a vector of weights for rejections.
Then, prediction $\widehat t$ is:
\[
\widehat t = \argmax_{t \in T' \cup R} \overline{w}_t,
\]
where no relation is predicted between $e_1$ and $e_2$ if $\widehat t \in R$.
This can be performed by simply adding negative examples (irrelevant relation mentions) to the dataset, and labelling them with a negative type $t'\in R$. This setup, however, creates a strong class imbalance (the negative class is often prevalent) and does not guarantee a rejection mechanism fit for new unknown relation types, as we require.
We describe 3 different rejection methods.

\begin{table*}[h!]
\small
\centering
\input{results/rejection_accuracy}
\caption{Rejection accuracy in the rejection pass.}
\label{tab:rejection_accuracy}
\end{table*}
\begin{table*}[h!]
\small
\centering
\input{results/f1_with_rejection}
\caption{Macro F1 measure in the retention pass, computed for (non reject) relation types.}
\label{tab:rejection_f1}
\end{table*}

\subsection{Proposed Methods}

\paragraph{Rejection Threshold.}
The simplest mechanism learns a threshold weight such that a relation type can be predicted if and only if its weight is higher than this threshold \citep{sabo2021revisiting}.
This threshold is initially set to 0.5 and further learned as a parameter of the model.

In this setting, $R = \{r\}$ is singleton,
and $u_r$ is a learned parameter.
It is easy to see that if $\forall t \in T': w_t < u_r$, the input relation is rejected.

\paragraph{Rejection Description.}
Another possible mechanism consists in assuming the relation class is defined as any other class, that is we use a type description like ``\texttt{There is no relation between the two entities.}'' \cite{sainz2021label,thomas2024one}.

In this setting, $R = \{ r \}$ and $u_r$ is computed as other relation type scores.

\paragraph{Rejection Prototypes.}
The last approach we test is a variant of \emph{multiple none of the above vectors} \citep{sabo2021revisiting}.
In this setting, $|R| > 1$,
and we learn several reject type prototypes (or rejection relation type representations).\footnote{This means that these vectors are learned parameters, and not outputs of a description encoder.}
Then, $u_r$ is the cosine similarity measure between the relation candidate representation and the reject type prototype for $r \in R$, computed in the same manner as other output weights.
We set $|R| = 5$, as the original article said this value had little impact on performance when ranging from 1 to 20.

\subsection{Training Loss}

Training a zero-shot RE model with a rejection mechanism is challenging.
First, every training instance is labeled with a training relation type.
Second, in the case of rejection prototypes, we have several rejection candidates to train.
Our loss is based on the squared hinge loss \citep{crammer2002svm}, defined as follows:\footnote{In preliminary experiments, we tested training with the negative log-likelihood loss, but it consistently results in very low performance.}
\[
\ell_\text{h2}(\overline{\vw} ; t) =
\max(0, 1 - \overline{w}_t + \max_{t' \neq t} \overline{w}_{t'})^2,
\]
where $\overline{\vw}$ is a vector of output scores and $t$ is the index of gold class.

We propose a novel loss for training with a rejection mechanism that is built around a \emph{ranking} objective:
the gold relation type should be ranked above all other relation types (pos.\ vs.\ neg.),
the gold relation type should be ranked above all rejection types (pos.\ vs.\ rej.),
and one rejection type should be ranked above all non-gold types (rej.\ vs.\ neg.).
Note that the last case is a partial labeling learning problem \citep{cour2011partial}: the target may be a non singleton set $R$.

Intuitively,
our approach learns to predict the gold relation type if it appears in the targeted types,
otherwise we should predict any rejection type.
This leads to the following aggregate loss:\footnote{Following all baseline implementations, $T'$ is the set of all gold classes that appear in the mini-batch.}
\begin{align*}
\ell_\text{rej.}(\overline{\vw} ; t)
=
&\phantom{+}\underbrace{\ell_\text{h2}\left(~
\left[ \overline w_t' \right]_{t' \in T'} ; t
~\right)}_{\text{pos.\ vs.\ neg.}}
\\
&+\underbrace{\ell_\text{h2}\left(~
\left[ \overline w_t' \right]_{t' \in R \cup \{t\}} ; t
~\right)}_{\text{pos.\ vs.\ rej.}}
\\
&+\underbrace{\min_{r \in R} \ell_\text{h2}\left(~
\left[ \overline w_t' \right]_{t' \in R \cup T' \setminus \{t\}} ; r
~\right)}_{\text{rej.\ vs.\ neg.}}.
\end{align*}
The last term can be understood as an ``hard EM'' approach for partial labeling \cite{corro-2024-fast}, where first search for the best rejection type, and then use it in a supervised loss.
Such partial labeling losses are also referred to as ``inf-losses'' \citep{pmlr-v119-cabannnes20a,stewart2023clustering}.

\subsection{Experiments}

\paragraph{Evaluation protocol.}
To simulate rejection cases in datasets which feature no explicit annotation for that, we perform two passes on the evaluation split:
\begin{enumerate}
    \item \textbf{Retention pass:} for each input, the model predicts one of the types in $T_\text{eval}$ (i.e.,\ model has to predict a relation type other than the rejection types in $R$);
    \item \textbf{Rejection pass:} each input has its gold output type removed from the target set (model has to predict rejection).
\end{enumerate}

\paragraph{Evaluation metrics.}
To measure the performance of each model,
we evaluate them with respect to two metrics:
the rejection accuracy in the rejection pass (Table~\ref{tab:rejection_accuracy}),
and the macro F1 measure on (non reject) relation types in the retention pass (Table~\ref{tab:rejection_f1}).

\paragraph{Analysis.}
Note that a model that learns to reject every input will have a rejection accuracy of $100\,\%$.
As such, when evaluating with a rejection mechanism, we search for a trade-off between the rejection accuracy and the F1 measure on relation types.
For example, \textsc{Emma} with the prototype rejection mechanism on \textsc{FewRel} learns to reject everything.

The threshold strategy performs worse over all models:
it pushes to learn either to reject almost everything or almost nothing.

\textsc{AlignRE} performs best overall using description and prototype rejection mechanisms:
its rejection accuracy is high while maintaining good F1 in the retention pass.
\textsc{RE-Matching} follows closely, especially on 10 or 15 unseen relation types.

%% file: results/rejection_accuracy.tex
\begin{tabular}{@{}l@{\hskip 4em}ccc@{\hskip 4em}ccc@{}}
\toprule
&\multicolumn{3}{@{}c@{\hskip 4em}}{\textsc{\textbf{FewRel}}}
&\multicolumn{3}{@{}c@{}}{\textsc{\textbf{WikiZSL}}}
\\
\cmidrule(lr{4em}){2-4}
\cmidrule{5-7}
& 5 & 10 & 15
& 5 & 10 & 15
\\
\midrule
\multicolumn{7}{l}{\textbf{\textsc{Emma}}}
\\
\midrule
Threshold
& \rstat{9.8}{4.0} & \rstat{7.3}{8.9} & \rstat{9.8}{9.3}
& \rstat{18.0}{6.4} & \rstat{28.7}{9.6} & \rstat{25.8}{12.7}
\\
Description
& \rstat{42.8}{16.6} & \rstat{42.5}{28.5} & \rstat{24.9}{15.3}
& \rstat{61.4}{14.2} & \rstat{49.8}{16.0} & \rstat{47.3}{9.5}
\\
Prototypes
& \rstat{100.0}{0.0} & \rstat{99.8}{0.3} & \rstat{100.0}{0.0}
& \rstat{37.1}{33.4} & \rstat{36.0}{54.5} & \rstat{67.5}{26.4}
\\
\midrule
\multicolumn{7}{l}{\textbf{\textsc{Re-Matching}}}
\\
\midrule
Threshold
& \rstat{100.0}{0.0} & \rstat{100.0}{0.0} & \rstat{100.0}{0.0}   
& \rstat{84.3}{16.1} & \rstat{97.6}{4.1} & \rstat{90.8}{14.8}
\\
Description
& \rstat{99.3}{0.6}  & \rstat{94.4}{7.1} & \rstat{86.9}{8.8}
& \rstat{91.8}{11.5} & \rstat{82.1}{18.4} & \rstat{85.7}{10.8}
\\
Prototypes
& \rstat{95.6}{6.4}  & \rstat{91.1}{6.3} & \rstat{81.5}{7.1}
& \rstat{98.8}{0.9} & \rstat{81.0}{12.1} & \rstat{73.5}{4.4}
\\
\midrule
\multicolumn{7}{l}{\textbf{\textsc{AlignRE}}}
\\
\midrule

Threshold
& \rstat{100.0}{0.0} & \rstat{99.7}{0.6} & \rstat{100.0}{0.0}
& \rstat{100.0}{0.0} & \rstat{99.7}{0.0} & \rstat{99.8}{0.1}
\\
Description
& \rstat{86.3}{10.3} &\rstat{26.6}{10.5} & \rstat{27.5}{4.3}
& \rstat{68.6}{14.0} & \rstat{34.3}{1.4} & \rstat{33.4}{3.9}
\\
Prototypes
& \rstat{99.1}{0.1} & \rstat{79.0}{10.0} & \rstat{65.5}{4.2}
& \rstat{94.9}{2.3} & \rstat{63.9}{10.5} & \rstat{69.7}{3.6}
\\
\bottomrule
\end{tabular}

%% file: results/f1_with_rejection.tex
\begin{tabular}{@{}l@{\hskip 4em}ccc@{\hskip 4em}ccc@{}}
\toprule
&\multicolumn{3}{@{}c@{\hskip 4em}}{\textsc{\textbf{FewRel}}}
&\multicolumn{3}{@{}c@{}}{\textsc{\textbf{WikiZSL}}}
\\
\cmidrule(lr{4em}){2-4}
\cmidrule{5-7}
& 5 & 10 & 15
& 5 & 10 & 15
\\
\midrule
\multicolumn{7}{l}{\textbf{\textsc{Emma}}}
\\
\midrule
Threshold
& \rstat{52.4}{1.9} & \rstat{31.8}{6.5} & \rstat{29.0}{11.0}
& \rstat{54.6}{6.0} & \rstat{47.1}{8.4} & \rstat{47.0}{8.1}
\\
Description
& \rstat{41.3}{12.2} & \rstat{30.1}{4.2} & \rstat{20.6}{6.0}
& \rstat{30.6}{8.5} & \rstat{24.8}{3.5} & \rstat{24.5}{6.0}
\\
Prototypes
& \rstat{0.0}{0.0} & \rstat{0.2}{0.3} & \rstat{0.0}{0.0}
& \rstat{29.9}{13.8} & \rstat{25.8}{19.2} & \rstat{23.7}{22.9}
\\
\midrule
\multicolumn{7}{l}{\textbf{\textsc{Re-Matching}}}
\\
\midrule
Threshold
& \rstat{0.1}{0.2} & \rstat{0.7}{1.2} & \rstat{0.1}{0.2}
& \rstat{26.7}{35.1} & \rstat{9.4}{3.1} & \rstat{26.7}{38.2}
\\
Description
& \rstat{30.8}{13.5} & \rstat{27.4}{19.7} & \rstat{24.7}{7.8}
& \rstat{49.1}{34.0} & \rstat{31.2}{7.4} & \rstat{29.1}{14.0}
\\
Prototypes
& \rstat{44.3}{10.9} & \rstat{35.9}{14.4} & \rstat{39.5}{11.6}
& \rstat{54.4}{1.1} & \rstat{28.9}{4.6} & \rstat{37.3}{12.5}
\\
\midrule
\multicolumn{7}{l}{\textbf{\textsc{AlignRE}}}
\\
\midrule

Threshold
& \rstat{0.0}{0.0} & \rstat{1.8}{3.0} & \rstat{0.0}{0.0}
& \rstat{0.1}{0.1} & \rstat{9.4}{0.1} & \rstat{4.2}{3.3}
\\
Description
& \rstat{79.7}{3.4} & \rstat{66.0}{2.5} & \rstat{55.9}{1.6}
& \rstat{64.5}{5.3} & \rstat{50.9}{8.3} & \rstat{51.1}{3.0}
\\
Prototypes
& \rstat{70.0}{8.0} & \rstat{71.9}{2.2} & \rstat{59.7}{1.3}
& \rstat{46.6}{3.7} & \rstat{34.5}{5.6} & \rstat{43.4}{6.8}

\\
\bottomrule
\end{tabular}

%% file: 6.conclusion.tex
\section{Conclusion}

In this work, we argued that most zero-shot RE works do not evaluate with real scenario in mind.
We therefore introduce a challenging use case,
and propose solutions to adapt several state-of-the-art models.

More specifically, we highlight two mandatory upgrades for efficient zero-shot RE:
single pass adaptation and rejection mechanism augmentation.
For both upgrades, we propose several strategies,
and evaluate them on two datasets.

%% file: biblio.bib
@inproceedings{levy2017zero,
	title        = {Zero-Shot Relation Extraction via Reading Comprehension},
	author       = {Levy, Omer  and Seo, Minjoon  and Choi, Eunsol  and Zettlemoyer, Luke},
	year         = 2017,
	month        = aug,
	booktitle    = {Proceedings of the 21st Conference on Computational Natural Language Learning ({C}o{NLL} 2017)},
	publisher    = {Association for Computational Linguistics},
	address      = {Vancouver, Canada},
	pages        = {333--342},
	doi          = {10.18653/v1/K17-1034},
	url          = {https://aclanthology.org/K17-1034/},
	editor       = {Levy, Roger  and Specia, Lucia}
}

@inproceedings{sainz2021label,
	title        = {Label Verbalization and Entailment for Effective Zero and Few-Shot Relation Extraction},
	author       = {Sainz, Oscar  and Lopez de Lacalle, Oier  and Labaka, Gorka  and Barrena, Ander  and Agirre, Eneko},
	year         = 2021,
	month        = nov,
	booktitle    = {Proceedings of the 2021 Conference on Empirical Methods in Natural Language Processing (EMNLP 2021)},
	publisher    = {Association for Computational Linguistics},
	address      = {Online and Punta Cana, Dominican Republic},
	pages        = {1199--1212},
	doi          = {10.18653/v1/2021.emnlp-main.92},
	url          = {https://aclanthology.org/2021.emnlp-main.92/},
	editor       = {Moens, Marie-Francine  and Huang, Xuanjing  and Specia, Lucia  and Yih, Scott Wen-tau}
}

@inproceedings{chen2021zs,
	title        = {{ZS}-{BERT}: Towards Zero-Shot Relation Extraction with Attribute Representation Learning},
	author       = {Chen, Chih-Yao  and Li, Cheng-Te},
	year         = 2021,
	month        = jun,
	booktitle    = {Proceedings of the 2021 Conference of the North American Chapter of the Association for Computational Linguistics: Human Language Technologies (NAACL 2021)},
	publisher    = {Association for Computational Linguistics},
	address      = {Online},
	pages        = {3470--3479},
	doi          = {10.18653/v1/2021.naacl-main.272},
	url          = {https://aclanthology.org/2021.naacl-main.272/},
	editor       = {Toutanova, Kristina  and Rumshisky, Anna  and Zettlemoyer, Luke  and Hakkani-Tur, Dilek  and Beltagy, Iz  and Bethard, Steven  and Cotterell, Ryan  and Chakraborty, Tanmoy  and Zhou, Yichao}
}

@inproceedings{wang2022rcl,
	title        = {{RCL}: Relation contrastive learning for zero-shot relation extraction},
	author       = {Wang, Shusen and Zhang, Bosen and Xu, Yajing and Wu, Yanan and Xiao, Bo},
	year         = 2022,
    month = jul,
	booktitle    = {Proceedings of Findings of the Association for Computational Linguistics: NAACL 2022},
	publisher    = {Association for Computational Linguistics},
    address={Seattle, Washington, USA},
	pages        = {2456--2468},
    doi = "10.18653/v1/2022.findings-naacl.188",
    url = "https://aclanthology.org/2022.findings-naacl.188/",
    editor = "Carpuat, Marine  and
      de Marneffe, Marie-Catherine  and
      Meza Ruiz, Ivan Vladimir"
}

@inproceedings{zhang2025custom,
	title        = {{CE-DA}: Custom Embedding and Dynamic Aggregation for Zero-Shot Relation Extraction},
	author       = {Zhang, Fu and Liu, He and Li, Zehan and Cheng, Jingwei},
	year         = 2025,
    month = jan,
	booktitle    = {Proceedings of the 31st International Conference on Computational Linguistics (COLING 2025)},
    address = "Abu Dhabi, United Arab Emirates",
    publisher = "Association for Computational Linguistics",
	pages        = {9814--9823},
    url = "https://aclanthology.org/2025.coling-main.656/",
    editor = "Rambow, Owen  and
      Wanner, Leo  and
      Apidianaki, Marianna  and
      Al-Khalifa, Hend  and
      Eugenio, Barbara Di  and
      Schockaert, Steven"
}

@inproceedings{li2024fusion,
	title        = {Fusion Makes Perfection: An Efficient Multi-Grained Matching Approach for Zero-Shot Relation Extraction},
	author       = {Li, Shilong  and Bai, Ge  and Zhang, Zhang  and Liu, Ying  and Lu, Chenji  and Guo, Daichi  and Liu, Ruifang  and Yong, Sun},
	year         = 2024,
	month        = jun,
	booktitle    = {Proceedings of the 2024 Conference of the North American Chapter of the Association for Computational Linguistics: Human Language Technologies (NAACL 2024)},
	publisher    = {Association for Computational Linguistics},
	address      = {Mexico City, Mexico},
	pages        = {79--85},
	doi          = {10.18653/v1/2024.naacl-short.7},
	url          = {https://aclanthology.org/2024.naacl-short.7/},
	editor       = {Duh, Kevin  and Gomez, Helena  and Bethard, Steven}
}

@inproceedings{zhao2023re,
	title        = {{RE}-{M}atching: A Fine-Grained Semantic Matching Method for Zero-Shot Relation Extraction},
	author       = {Zhao, Jun  and Zhan, WenYu  and Zhao, Xin  and Zhang, Qi  and Gui, Tao  and Wei, Zhongyu  and Wang, Junzhe  and Peng, Minlong  and Sun, Mingming},
	year         = 2023,
	month        = jul,
	booktitle    = {Proceedings of the 61st Annual Meeting of the Association for Computational Linguistics (ACL 2023)},
	publisher    = {Association for Computational Linguistics},
	address      = {Toronto, Canada},
	pages        = {6680--6691},
	doi          = {10.18653/v1/2023.acl-long.369},
	url          = {https://aclanthology.org/2023.acl-long.369/},
	editor       = {Rogers, Anna  and Boyd-Graber, Jordan  and Okazaki, Naoaki}
}

@article{yang2024cl,
	title        = {{CL}\&{CD}: Contrastive Learning and Cluster Description for Zero-Shot Relation Extraction},
	author       = {Yang, Zongqiang and Fei, Junbo and Tan, Zhen and Tang, Jiuyang and Zhao, Xiang},
	year         = 2024,
	journal      = {Knowledge-Based Systems},
	publisher    = {Elsevier},
	volume       = 293,
	pages        = 111652,
    doi = {10.1016/j.knosys.2024.111652}
}

@inproceedings{wang2022matchprompt,
	title        = {Match{P}rompt: Prompt-based open relation extraction with semantic consistency guided clustering},
	author       = {Wang, Jiaxin and Zhang, Lingling and Liu, Jun and Liang, Xi and Zhong, Yujie and Wu, Yaqiang},
	year         = 2022,
    month = dec,
	booktitle    = {Proceedings of the 2022 Conference on Empirical Methods in Natural Language Processing (EMNLP 2022)},
    publisher = "Association for Computational Linguistics",
    address = "Abu Dhabi, United Arab Emirates",
	pages        = {7875--7888},
    doi = "10.18653/v1/2022.emnlp-main.537",
    url = "https://aclanthology.org/2022.emnlp-main.537/",
    editor = "Goldberg, Yoav  and
      Kozareva, Zornitsa  and
      Zhang, Yue"
}

@inproceedings{li2024alignre,
	title        = {Align{RE}: An Encoding and Semantic Alignment Approach for Zero-Shot Relation Extraction},
	author       = {Li, Zehan and Zhang, Fu and Cheng, Jingwei},
	year         = 2024,
    month = aug,
	booktitle    = {Proceedings of Findings of the Association for Computational Linguistics: ACL 2024},
    publisher = "Association for Computational Linguistics",
    address = "Bangkok, Thailand",
	pages        = {2957--2966},
    doi = "10.18653/v1/2024.findings-acl.174",
    url = "https://aclanthology.org/2024.findings-acl.174/",
    editor = "Ku, Lun-Wei  and
      Martins, Andre  and
      Srikumar, Vivek",
}

@inproceedings{chia2022relationprompt,
	title        = {{R}elation{P}rompt: Leveraging Prompts to Generate Synthetic Data for Zero-Shot Relation Triplet Extraction},
	author       = {Chia, Yew Ken  and Bing, Lidong  and Poria, Soujanya  and Si, Luo},
	year         = 2022,
	month        = may,
	booktitle    = {Proceedings of Findings of the Association for Computational Linguistics: ACL 2022},
	publisher    = {Association for Computational Linguistics},
	address      = {Dublin, Ireland},
	pages        = {45--57},
	doi          = {10.18653/v1/2022.findings-acl.5},
	url          = {https://aclanthology.org/2022.findings-acl.5/},
	editor       = {Muresan, Smaranda  and Nakov, Preslav  and Villavicencio, Aline}
}

@inproceedings{zhang2023aligning,
    title   = "Aligning Instruction Tasks Unlocks Large Language Models as Zero-Shot Relation Extractors",
    author = "Zhang, Kai  and
      Jimenez Gutierrez, Bernal  and
      Su, Yu",
    year    = 2023,
    month   = jul,
    
    booktitle   = "Findings of the Association for Computational Linguistics: ACL 2023",
    publisher   = "Association for Computational Linguistics",
    address = "Toronto, Canada",
    pages   = "794--812",
    doi = "10.18653/v1/2023.findings-acl.50",
    url = "https://aclanthology.org/2023.findings-acl.50/",    
    editor  = "Rogers, Anna  and
      Boyd-Graber, Jordan  and
      Okazaki, Naoaki",
}

@inproceedings{naguib2024fewshot,
    title   = "Few-shot clinical entity recognition in {E}nglish, {F}rench and {S}panish: masked language models outperform generative model prompting",
    author  = "Naguib, Marco  and
      Tannier, Xavier  and
      N{\'e}v{\'e}ol, Aur{\'e}lie",
    year    = 2024,
    month   = nov,
    booktitle   = "Findings of the Association for Computational Linguistics: EMNLP 2024",
    publisher   = "Association for Computational Linguistics",
    address = "Miami, Florida, USA",
    pages   = "6829--6852",
    doi = "10.18653/v1/2024.findings-emnlp.400",
    url = "https://aclanthology.org/2024.findings-emnlp.400/",
    editor  = "Al-Onaizan, Yaser  and
      Bansal, Mohit  and
      Chen, Yun-Nung",
}

@inproceedings{boylan2025glirel,
	title        = {{GL}i{REL} - {G}eneralist Model for Zero-Shot Relation Extraction},
	author       = {Boylan, Jack  and Hokamp, Chris  and Ghalandari, Demian Gholipour},
	year         = 2025,
	month        = apr,
	booktitle    = {Proceedings of the 2025 Conference of the Nations of the Americas Chapter of the Association for Computational Linguistics: Human Language Technologies (NAACL 2025)},
	publisher    = {Association for Computational Linguistics},
	address      = {Albuquerque, New Mexico, USA},
	pages        = {8230--8245},
	doi          = {10.18653/v1/2025.naacl-long.418},
	optisbn         = {979-8-89176-189-6},
	url          = {https://aclanthology.org/2025.naacl-long.418/},
	editor       = {Chiruzzo, Luis  and Ritter, Alan  and Wang, Lu}
}

@article{sabo2021revisiting,
	title        = {Revisiting few-shot relation classification: Evaluation data and classification schemes},
	author       = {Sabo, Ofer and Elazar, Yanai and Goldberg, Yoav and Dagan, Ido},
    year = "2021",
    journal = "Transactions of the Association for Computational Linguistics",
    publisher = "MIT Press",
    optaddress = "Cambridge, Massachusetts, USA",
    volume = "9",
    pages = "691--706",
    doi = "10.1162/tacl_a_00392",
    url = "https://aclanthology.org/2021.tacl-1.42/",
	editor = "Roark, Brian  and
      Nenkova, Ani",
}

@inproceedings{han2018fewrel,
	title        = {{F}ew{R}el: A Large-Scale Supervised Few-Shot Relation Classification Dataset with State-of-the-Art Evaluation},
	author       = {Han, Xu  and Zhu, Hao  and Yu, Pengfei  and Wang, Ziyun  and Yao, Yuan  and Liu, Zhiyuan  and Sun, Maosong},
	year         = 2018,
	month        = {oct},
	booktitle    = {Proceedings of the 2018 Conference on Empirical Methods in Natural Language Processing (EMNLP 2018)},
	publisher    = {Association for Computational Linguistics},
	address      = {Brussels, Belgium},
	pages        = {4803--4809},
	doi          = {10.18653/v1/D18-1514},
	url          = {https://aclanthology.org/D18-1514/},
	editor       = {Riloff, Ellen  and Chiang, David  and Hockenmaier, Julia  and Tsujii, Jun{'}ichi}
}

@article{habibi2017deep,
	title        = {Deep learning with word embeddings improves biomedical named entity recognition},
	author       = {Habibi, Maryam and Weber, Leon and Neves, Mariana and Wiegandt, David Luis and Leser, Ulf},
	year         = 2017,
	journal      = {Bioinformatics},
	publisher    = {Oxford University Press},
	volume       = 33,
	number       = 14,
	pages        = {i37--i48},
    doi={10.1093/bioinformatics/btx228},
    url = {https://doi.org/10.1093/bioinformatics/btx228}
}

@inproceedings{hamad2024fire,
	title        = {{FIRE}: A Dataset for FInancial Relation Extraction},
	author       = {Hamad, Hassan and Thakur, Abhinav Kumar and Kolleri, Nijil and Pulikodan, Sujith and Chugg, Keith},
    year = "2024",
    month = jun,
    booktitle = "Proceedings of Findings of the Association for Computational Linguistics: NAACL 2024",
    publisher = "Association for Computational Linguistics",
    address = "Mexico City, Mexico",
    pages = "3628--3642",
    doi = "10.18653/v1/2024.findings-naacl.230",
    url = "https://aclanthology.org/2024.findings-naacl.230/",
	editor = "Duh, Kevin  and
      Gomez, Helena  and
      Bethard, Steven",
}

@inproceedings{zhong2020does,
	title        = {How Does {NLP} Benefit Legal System: A Summary of Legal Artificial Intelligence},
	author       = {Zhong, Haoxi  and Xiao, Chaojun  and Tu, Cunchao  and Zhang, Tianyang  and Liu, Zhiyuan  and Sun, Maosong},
	year         = 2020,
	month        = jul,
	booktitle    = {Proceedings of the 58th Annual Meeting of the Association for Computational Linguistics (ACL 2020)},
	publisher    = {Association for Computational Linguistics},
	address      = {Online},
	pages        = {5218--5230},
	doi          = {10.18653/v1/2020.acl-main.466},
	url          = {https://aclanthology.org/2020.acl-main.466/},
	editor       = {Jurafsky, Dan and Chai, Joyce and Schluter, Natalie and Tetreault, Joel}
}

@article{hendrickx2024machine,
	title        = {Machine learning with a reject option: A survey},
	author       = {Hendrickx, Kilian and Perini, Lorenzo and Van der Plas, Dries and Meert, Wannes and Davis, Jesse},
	year         = 2024,
	journal      = {Machine Learning},
	publisher    = {Springer},
	volume       = 113,
	number       = 5,
	pages        = {3073--3110},
    doi={10.1007/s10994-024-06534-x},
}

@inproceedings{hendrycks2016baseline,
	title        = {A baseline for detecting misclassified and out-of-distribution examples in neural networks},
	author       = {Hendrycks, Dan and Gimpel, Kevin},
	year         = 2017,
    month={apr},
	booktitle      = {Proceedings of the 5th International Conference on Learning Representations (ICLR 2017)},
    publisher={ICLR},
    address={Toulon, France},
    pages = {2410--2421},
    doi={10.48550/arXiv.1610.02136},
    editor = {Curran Associates, Inc.}
}

@article{barandas2022uncertainty,
	title        = {Uncertainty-based rejection in machine learning: Implications for model development and interpretability},
	author       = {Barandas, Mar{\'\i}lia and Folgado, Duarte and Santos, Ricardo and Sim{\~a}o, Raquel and Gamboa, Hugo},
	year         = 2022,
	journal      = {Electronics},
	publisher    = {MDPI},
	volume       = 11,
	number       = 3,
	pages        = 396,
    optarticle-number={396},
    doi={10.3390/electronics11030396},
    url = {https://www.mdpi.com/2079-9292/11/3/396},
}

@inproceedings{soares2019matching,
	title        = {Matching the Blanks: Distributional Similarity for Relation Learning},
	author       = {Baldini Soares, Livio  and FitzGerald, Nicholas  and Ling, Jeffrey  and Kwiatkowski, Tom},
	year         = 2019,
	month        = jul,
	booktitle    = {Proceedings of the 57th Annual Meeting of the Association for Computational Linguistics (ACL 2019)},
	publisher    = {Association for Computational Linguistics},
	address      = {Florence, Italy},
	pages        = {2895--2905},
	doi          = {10.18653/v1/P19-1279},
	url          = {https://aclanthology.org/P19-1279/},
	editor       = {Korhonen, Anna  and Traum, David  and M{\`a}rquez, Llu{\'i}s}
}

@inproceedings{gravier2023derriere,
	title        = {Derri{\`e}re les plongements de relations},
	author       = {Thomas, Hugo and Gravier, Guillaume and S{\'e}billot, Pascale},
	year         = 2023,
    month = jun,
    booktitle = {Proceedings of the 30e Conf{\'e}rence sur le Traitement Automatique des Langues Naturelles (TALN 2023)},
    publisher = "ATALA",
    address = "Paris, France",
    pages = "311--322",
    url = "https://aclanthology.org/2023.jeptalnrecital-long.24/",
    editor = "Servan, Christophe  and Vilnat, Anne",
}

@article{thomas2024one,
	title        = {One-shot relation retrieval in news archives: adapting N-way K-shot relation classification for efficient knowledge extraction},
	author       = {Thomas, Hugo and Gravier, Guillaume and S{\'e}billot, Pascale},
	year         = 2024,
    month = sep,
    optbooktitle={Proceedings of the 28th KES International Conference on Knowledge-Based and Intelligent Information \& Engineering Systems (KES 2024)},
	journal      = {Procedia Computer Science},
	publisher    = {Elsevier},
    address = {Sevilla, Spain},
	volume       = 246,
	pages        = {1060--1069},
    doi={10.1016/j.procs.2024.09.525},
    editor={Toro, Carlos and Rios, Sebastian A., and Howlett, Robert J. and Jain, Lakhmi C.},
}

@article{yang2024survey,
	title        = {A Survey of Information Extraction Based on Deep Learning},
	author       = {Yang, Yang and Wu, Zhilei and Yang, Yuexiang and Lian, Shuangshuang and Guo, Fengjie and Wang, Zhiwei},
	year         = 2022,
	journal      = {Applied Sciences},
    publisher={MDPI},
	volume       = 12,
	number       = 19,
    pages = 9691,
	optarticle-number = 9691,
	doi          = {10.3390/app12199691},
	url          = {https://www.mdpi.com/2076-3417/12/19/9691},
	optissn         = {2076-3417},
}

@inproceedings{liu2024unleashing,
    title   = "Unleashing the Power of Large Language Models in Zero-shot Relation Extraction via Self-Prompting",
    author  = "Liu, Siyi  and
      Li, Yang  and
      Li, Jiang  and
      Yang, Shan  and
      Lan, Yunshi",
    year    = 2024,
    month   = nov,
    booktitle   = "Findings of the Association for Computational Linguistics: EMNLP 2024",
    publisher   = "Association for Computational Linguistics",
    address = "Miami, Florida, USA",
    pages   = "13147--13161",
    doi = "10.18653/v1/2024.findings-emnlp.769",
    url = "https://aclanthology.org/2024.findings-emnlp.769/",
    editor = "Al-Onaizan, Yaser  and Bansal, Mohit  and Chen, Yun-Nung",
}

@inproceedings{chen2019uhop,
	title        = {{U}{H}op: An Unrestricted-Hop Relation Extraction Framework for Knowledge-Based Question Answering},
	author       = {Chen, Zi-Yuan and Chang, Chih-Hung and Chen, Yi-Pei and Nayak, Jijnasa and Ku, Lun-Wei},
	year         = 2019,
    month = jun,
    booktitle = "Proceedings of the 2019 Conference of the North {A}merican Chapter of the Association for Computational Linguistics: Human Language Technologies (NAACL 2019)",
    publisher = "Association for Computational Linguistics",
    address = "Minneapolis, Minnesota, USA",
	pages        = {345--356},
    doi = "10.18653/v1/N19-1031",
    url = "https://aclanthology.org/N19-1031/",
    editor = "Burstein, Jill and Doran, Christy  and Solorio, Thamar",
}

@inproceedings{agosti2018relation,
	title        = {A Relation Extraction Approach for Clinical Decision Support},
	author       = {Agosti, Maristella and Di Nunzio, Giorgio and Marchesin, Stefano and Silvello, Gianmaria and others},
	year         = 2018,
    month=oct,
	booktitle    = {Proceedings of the CIKM 2018 Workshops co-located with 27th ACM International Conference on Information and Knowledge Management (CIKM 2018); 12th International Workshop on Data and Text Mining in Biomedical Informatics (DTMBio 2018)}, 
    publisher={CEUR Workshop Proceedings},
    address={Torino, Italy},
	volume       = 2482,
    url={https://ceur-ws.org/Vol-2482/},
	editors = {Alfredo Cuzzocrea and Francesco Bonchi and Dimitrios Gunopulos}
}

@article{hang2025few,
  title={Few-Shot Relation Extraction Based on Prompt Learning: A Taxonomy, Survey, Challenges and Future Directions},
  author={Hang, Tingting and Liu, Shuting and Feng, Jun and Djigal, Hamza and Huang, Jun},
  year={2025},
  journal={ACM Computing Surveys},
  publisher={ACM},
  volume=58,
  number=2,
  pages={40},
  optarticle-number={40},
  doi={10.1145/3746281}
}

@article{zhao2024comprehensive,
  title={A comprehensive survey on relation extraction: Recent advances and new frontiers},
  author={Zhao, Xiaoyan and Deng, Yang and Yang, Min and Wang, Lingzhi and Zhang, Rui and Cheng, Hong and Lam, Wai and Shen, Ying and Xu, Ruifeng},
  year={2024},
  journal={ACM Computing Surveys},
  publisher={ACM},
  volume={56},
  number={11},
  pages={293},
  optarticle-number={293},
  doi={10.1145/3674501}
}

@inproceedings{pai-etal-2024-survey,
    title = "A Survey on Open Information Extraction from Rule-based Model to Large Language Model",
    author = "Pai, Liu  and
      Gao, Wenyang  and
      Dong, Wenjie  and
      Ai, Lin  and
      Gong, Ziwei  and
      Huang, Songfang  and
      Zongsheng, Li  and
      Hoque, Ehsan  and
      Hirschberg, Julia  and
      Zhang, Yue",
    year = "2024",
    month = "nov",
    booktitle = "Proceedings of Findings of the Association for Computational Linguistics: EMNLP 2024",
    publisher = "Association for Computational Linguistics",
    address = "Miami, Florida, USA",
    pages = "9586--9608",
    doi = "10.18653/v1/2024.findings-emnlp.560",
    url = "https://aclanthology.org/2024.findings-emnlp.560/",
    editor = "Al-Onaizan, Yaser  and
      Bansal, Mohit  and
      Chen, Yun-Nung",
}

@inproceedings{deng2024information,
  title={Information extraction in low-resource scenarios: Survey and perspective},
  author={Deng, Shumin and Ma, Yubo and Zhang, Ningyu and Cao, Yixin and Hooi, Bryan},
  year={2024},
  month="dec",
  booktitle={Proceedings of the 15th IEEE International Conference on Knowledge Graphs (ICKG 2024)},
  publisher={IEEE},
  address={Abu Dhabi, United Arab Emirates},
  pages={33--49},
  doi={10.48550/arXiv.2202.08063},
  editor = {Chen, Huajun and Fensel, Anna and Zhu, Xingquan (Hill) and Wattenhofer, Roger and Wu, Xindong},
}

@inproceedings{stoica2020re,
  title={Re-{TACRED}: A new relation extraction dataset},
  author={Stoica, George and Platanios, Emmanouil Antonios and P{\'o}czos, Barnab{\'a}s},
  year={2020},
  month=dec,
  booktitle={Proceedings of the 4th Knowledge Representation and Reasoning Meets Machine Learning Workshop (KR2ML 2020, at NeurIPS'20)},
  address={Online},
  url={https://kr2ml.github.io/2020/papers/KR2ML_12_paper.pdf},
}

@InProceedings{riedel2010nyt,
  title="Modeling Relations and Their Mentions without Labeled Text",
  author="Riedel, Sebastian
and Yao, Limin
and McCallum, Andrew",
  year="2010",
  month="sep",
  booktitle="Proceedings of the 2010 European Conference on Machine Learning and Knowledge Discovery in Databases (ECML PKDD'10)",
  publisher="Springer-Verlag",
  address="Barcelona, Spain",
  pages="148--163",
  doi={doi/10.5555/1888339.1888350},
  optisbn="978-3-642-15939-8",
  editor="Balc{\'a}zar, Jos{\'e} Luis
and Bonchi, Francesco
and Gionis, Aristides
and Sebag, Mich{\`e}le",
}

@inproceedings{li2023revisiting,
  title={Revisiting Large Language Models as Zero-shot Relation Extractors},
  author={Li, Guozheng and Wang, Peng and Ke, Wenjun},
  year = "2023",
  month = dec,
  booktitle={Proceeding of Findings of the Association for Computational Linguistics: EMNLP 2023},
  publisher = "Association for Computational Linguistics",
  address = "Singapore",
  pages={6877--6892},
  url = "https://aclanthology.org/2023.findings-emnlp.459/",
  editor = "Bouamor, Houda  and
      Pino, Juan  and
      Bali, Kalika",
}

@inproceedings{sarrouti2022comparing,
  title={Comparing encoder-only and encoder-decoder transformers for relation extraction from biomedical texts: An empirical study on ten benchmark datasets},
  author={Sarrouti, Mourad and Tao, Carson and Randriamihaja, Yoann Mamy},
  year={2022},
  month = may,
  booktitle={Proceedings of the 21st Workshop on Biomedical Language Processing (BioNLP 2021)},
  publisher = "Association for Computational Linguistics",
  address = "Dublin, Ireland",
  pages={376--382},
  doi = "10.18653/v1/2022.bionlp-1.37",
  url = "https://aclanthology.org/2022.bionlp-1.37/",
  editor = "Demner-Fushman, Dina  and
      Cohen, Kevin Bretonnel  and
      Ananiadou, Sophia  and
      Tsujii, Junichi",
}

@inproceedings{khattab2020colbert,
    title   = {ColBERT: Efficient and Effective Passage Search via Contextualized Late Interaction over {BERT}},
    author  = {Khattab, Omar and Zaharia, Matei},
    year    = {2020},
    month   = jul,
    booktitle   = {Proceedings of the 43rd International ACM SIGIR Conference on Research and Development in Information Retrieval (SIGIR 2020)},
    publisher   = {Association for Computing Machinery},
    address = {New York, NY, USA},
    pages   = {39–48},
    doi = {10.1145/3397271.3401075},
    url = {https://doi.org/10.1145/3397271.3401075},
    editor  = {Huang, Jimmy X. and Chang, Yi}
}

@inproceedings{gao2019fewrel,
    title   = "{F}ew{R}el 2.0: Towards More Challenging Few-Shot Relation Classification",
    author  = "Gao, Tianyu  and
      Han, Xu  and
      Zhu, Hao  and
      Liu, Zhiyuan  and
      Li, Peng  and
      Sun, Maosong  and
      Zhou, Jie",
    year    = 2019,
    month   = nov,
    booktitle   = "Proceedings of the 2019 Conference on Empirical Methods in Natural Language Processing and the 9th International Joint Conference on Natural Language Processing (EMNLP-IJCNLP 2019)",
    publisher   = "Association for Computational Linguistics",
    address = "Hong Kong, China",
    pages   = "6250--6255",
    doi = "10.18653/v1/D19-1649",
    url = "https://aclanthology.org/D19-1649/",  
    editor  = "Inui, Kentaro  and
      Jiang, Jing  and
      Ng, Vincent  and
      Wan, Xiaojun",
}

@inproceedings{bassignana2022mean,
    title = "What Do You Mean by Relation Extraction? {A} Survey on Datasets and Study on Scientific Relation Classification",
    author = "Bassignana, Elisa  and
      Plank, Barbara",
    year = "2022",    
    month = may,
    booktitle = "Proceedings of the 60th Annual Meeting of the Association for Computational Linguistics: Student Research Workshop (ACL 2022)",
    publisher = "Association for Computational Linguistics",
    address = "Dublin, Ireland",
    pages = "67--83",
    doi = "10.18653/v1/2022.acl-srw.7",
    url = "https://aclanthology.org/2022.acl-srw.7/",
    editor = "Louvan, Samuel  and
      Madotto, Andrea  and
      Madureira, Brielen",
}

@inproceedings{devlin-etal-2019-bert,
    title = "{BERT}: Pre-training of Deep Bidirectional Transformers for Language Understanding",
    author = "Devlin, Jacob  and
      Chang, Ming-Wei  and
      Lee, Kenton  and
      Toutanova, Kristina",
    editor = "Burstein, Jill  and
      Doran, Christy  and
      Solorio, Thamar",
    booktitle = "Proceedings of the 2019 Conference of the North {A}merican Chapter of the Association for Computational Linguistics: Human Language Technologies (NAACL 2019)",
    month = jun,
    year = "2019",
    address = "Minneapolis, Minnesota, USA",
    publisher = "Association for Computational Linguistics",
    url = "https://aclanthology.org/N19-1423/",
    doi = "10.18653/v1/N19-1423",
    pages = "4171--4186"
}

@inproceedings{reimers-gurevych-2019-sentence,
    title = "Sentence-{BERT}: Sentence Embeddings using {S}iamese {BERT}-Networks",
    author = "Reimers, Nils  and
      Gurevych, Iryna",
    editor = "Inui, Kentaro  and
      Jiang, Jing  and
      Ng, Vincent  and
      Wan, Xiaojun",
    booktitle = "Proceedings of the 2019 Conference on Empirical Methods in Natural Language Processing and the 9th International Joint Conference on Natural Language Processing (EMNLP-IJCNLP)",
    month = nov,
    year = "2019",
    address = "Hong Kong, China",
    publisher = "Association for Computational Linguistics",
    url = "https://aclanthology.org/D19-1410/",
    doi = "10.18653/v1/D19-1410",
    pages = "3982--3992"
}

@inproceedings{
loshchilov2018decoupled,
title={Decoupled Weight Decay Regularization},
author={Ilya Loshchilov and Frank Hutter},
booktitle      = {Proceedings of the 7th International Conference on Learning Representations (ICLR 2019)},
publisher={OpenReview.net},
    address={New Orleans, Lousiana, USA},
    optpages = {},
notes={13 pages},
year={2019},
url = {https://openreview.net/forum?id=Bkg6RiCqY7}
}

@article{crammer2002svm,
author = {Crammer, Koby and Singer, Yoram},
title = {On the algorithmic implementation of multiclass kernel-based vector machines},
year = {2002},
issue_date = {3/1/2002},
publisher = {JMLR.org},
volume = {2},
issn = {1532-4435},
journal = {Journal of Machine Learning Research},
month = mar,
pages = {265--292},
numpages = {28},
url={https://www.jmlr.org/papers/volume2/crammer01a/crammer01a.pdf}
}

@InProceedings{pmlr-v119-cabannnes20a,
  title = 	 {Structured Prediction with Partial Labelling through the Infimum Loss},
  author =       {Cabannes, Vivien and Rudi, Alessandro and Bach, Francis},
  booktitle = 	 {Proceedings of the 37th International Conference on Machine Learning (ICML 2020)},
  pages = 	 {1230--1239},
  year = 	 {2020},
  editor = 	 {III, Hal Daumé and Singh, Aarti},
  optvolume = 	 {119},
  optseries = 	 {Proceedings of Machine Learning Research},
  optmonth = 	 {13--18 Jul},
  publisher =    {PMLR},
  url = 	 {https://proceedings.mlr.press/v119/cabannnes20a.html}
}

@inproceedings{stewart2023clustering,
 author = {Stewart, Lawrence and Bach, Francis and Llinares-Lopez, Felipe and Berthet, Quentin},
 title = {Differentiable Clustering with Perturbed Spanning Forests},
 booktitle = {Advances in Neural Information Processing Systems (NeurIPS 2023)},
 editor = {A. Oh and T. Naumann and A. Globerson and K. Saenko and M. Hardt and S. Levine},
 pages = {31158--31176},
 publisher = {Curran Associates, Inc.},
 url = {https://proceedings.neurips.cc/paper_files/paper/2023/file/637a456d89289769ac1ab29617ef7213-Paper-Conference.pdf},
 volume = {36},
 year = {2023}
}

@inproceedings{corro-2024-fast,
    title = "A Fast and Sound Tagging Method for Discontinuous Named-Entity Recognition",
    author = "Corro, Caio",
    editor = "Al-Onaizan, Yaser  and
      Bansal, Mohit  and
      Chen, Yun-Nung",
    booktitle = "Proceedings of the 2024 Conference on Empirical Methods in Natural Language Processing (EMNLP 2024)",
    month = nov,
    year = "2024",
    address = "Miami, Florida, USA",
    publisher = "Association for Computational Linguistics",
    url = "https://aclanthology.org/2024.emnlp-main.1087/",
    doi = "10.18653/v1/2024.emnlp-main.1087",
    pages = "19506--19518"
}

@article{cour2011partial,
  author  = {Timothee Cour and Ben Sapp and Ben Taskar},
  title   = {Learning from Partial Labels},
  journal = {Journal of Machine Learning Research},
  year    = {2011},
  volume  = {12},
  number  = {42},
  pages   = {1501--1536},
  url     = {http://jmlr.org/papers/v12/cour11a.html}
}
